# RUMC: A Rule-based Classifier Inspired by Evolutionary Methods

Melvin Mokhtari
melvim1@mcmaster.ca
Department of Computing and Software
McMaster University
Hamilton, CA

## ABSTRACT
As the field of data analysis grows rapidly due to the large amounts of data being generated, effective data classification has become increasingly important. This paper introduces the RUle Mutation Classifier (RUMC), which represents a significant improvement over the Rule Aggregation ClassifiER (RACER). RUMC uses innovative rule mutation techniques based on evolutionary methods to improve classification accuracy. In tests with forty datasets from OpenML and the UCI Machine Learning Repository, RUMC consistently outperformed twenty other well-known classifiers, demonstrating its ability to uncover valuable insights from complex data.

## 1 INTRODUCTION

Technology has rooted itself in every corner of our everyday lives, leading to a remarkable increase in the amount of data generated. This surge has made data classification algorithms important, as they help organize and extract meaningful information [29]. These algorithms play a vital role in decision-making processes and significantly influence machine learning systems, aiding in the management of information. They are applied in various fields, including healthcare [8, 18, 38], finance [26, 43], agriculture [14], and education [35, 42], reflecting the growing necessity for precise classification tools [10].

Data classification methods can be grouped into three main types: eager learners, lazy learners, and other classification strategies [32]. Eager learners, such as decision trees (like CART, C4.5 [34], and LMT, along with newer ones such as CHAID* [20], ForestPA [4], ForEx++ [5], and SPAARC [45]), Bayesian classifiers (such as Naïve Bayes), rule-based systems (including ROPAC-M [28], ROPAC-L [28], RACER, JRip, and PART), support vector machines (like SMO in SVM), and neural networks (like MLP), build a model from the training data before making any predictions. On the other hand, lazy learners, like the K-Nearest Neighbor (KNN) method, delay the modeling process until a new instance requires classification. Alternative classification strategies encompass fuzzy logic and rough set theory, providing diverse approaches to the task. This category also includes ensemble methods such as Bagging, Random Forest, CSForest [36], and Optimized Forest [3], which utilize multiple models to enhance prediction accuracy.

In this context, rule-based classification algorithms have gained prominence due to their straightforward nature, interpretability, and transparency in decision-making [22, 39]. These algorithms create rules in an "if-then" format, where the conditions on the input features determine the assigned class labels, making them particularly useful in areas where clarity is essential [25].

The Rule Aggregating ClassifiER (RACER) [7] is a rule-based classification algorithm that generates initial rules from training dataset records with the same mechanism. However, these rules tend to be too specific, making them less effective for classifying new data, particularly when working with small datasets that have few distinct instances. To address this challenge, I introduce the RUle Mutation Classifier (RUMC), a novel algorithm that enhances the capabilities of RACER. RUMC aims to improve the handling of various datasets, including high-dimensional and low-sample-size data, by incorporating a new rule mutation technique, which is to iteratively refine the current initial rule set until a better fitness is reached. The primary objective is to achieve higher accuracy than RACER and also to surpass other famous classifiers in performance across both small and large datasets.

The organization of this paper is structured as follows: Section 2 reviews relevant literature. The RACER algorithm is detailed in Section 3. In Section 4, I provide an in-depth explanation of RUMC. Section 5 discusses a performance evaluation using benchmark datasets and presents the experimental results on classification accuracy. Finally, Section 6 concludes the paper with a discussion of the findings and potential future research directions.

## 2 RELATED WORK

Rule-based classifiers are a specific type of data classification algorithm that derive a set of rules from a training dataset to categorize new instances. Unlike decision tree classifiers, which follow a hierarchical structure, rule-based classifiers offer more flexible representations [31]. Notable examples include PRISM [11], one of the earliest rule-based classifiers, which generates rules from all examples of a particular class. IREP combines reduced error pruning with a divide-and-conquer strategy to minimize classification errors [16]. Similarly, RIPPER [12] builds rules iteratively until a stopping criterion is met, while PART [1] learns one rule at a time without requiring global optimization, generating rules repeatedly. This overview highlights the foundational algorithms in this field. Recently, there has been growing interest in optimizing rule extraction and enhancing the performance of these classifiers.

One method introduced for binary classification involves partitioning each class into smaller subtables to improve accuracy [19]. Another approach utilizes particle swarm optimization with quantum qubit operations to develop fuzzy rule-based classifiers, focusing on computational efficiency and accuracy [27]. Research has also examined the complexities associated with rule ordering, emphasizing the NP-hard nature of certain optimization processes [15].

In 2019, a novel annealing strategy was introduced to improve a rule pruning technique within an Ant Colony Optimization-based classification system, addressing the challenge of generating overly complex rules [6]. Another study focused on rule induction through iterated local search to generate effective classification rules that uncover underlying data patterns [21].

Other recent rule-based algorithms include eRules, which manages streaming data with a sliding window to quickly correct misclassifications [37]. Its updated version, G-eRules, addresses real-time rule induction [24]. The RRULES algorithm [31] enhances the original RULES algorithm [33] by refining rule extraction and emphasizing the verification of stopping conditions.

In 2019, the RACER algorithm was introduced as a high-performance rule-based classifier, notable for its unique rule representation and combination techniques [7]. Research subsequently explored various discretization methods to enhance RACER's accuracy and comprehensibility, with the MDLP method showing improved performance [40]. In 2024, the Rule OPtimized Aggregation Classifier (ROPAC) [28] was developed as an extension of RACER to improve accuracy across diverse data environments, available in ROPAC-M and ROPAC-L forms. This new approach incorporates an optimization step that analyzes the frequency of zero bits in the rules, allowing for the identification and enhancement of less effective rules by flipping bits to create a more optimized rule set.

This paper presents the RUle Mutation Classifier (RUMC), an extension of RACER designed to improve overall accuracy in diverse data environments while addressing its existing limitations through a mutating phase inspired by the evolutionary approaches.

## 3 STATE-OF-THE-ART CLASSIFIER: RACER

The RACER converts each training record into an initial rule and subsequently merges these to create more generalized rules, offering a wide range of options for rule merging [7]. Unlike many advanced classifiers that select features independently [23], RACER utilizes all attributes simultaneously, resulting in quicker and more accurate classifications. This effective approach, makes RACER a highly efficient and reliable choice for modern data classification. Figure 1 illustrates the workflow of this classifier.

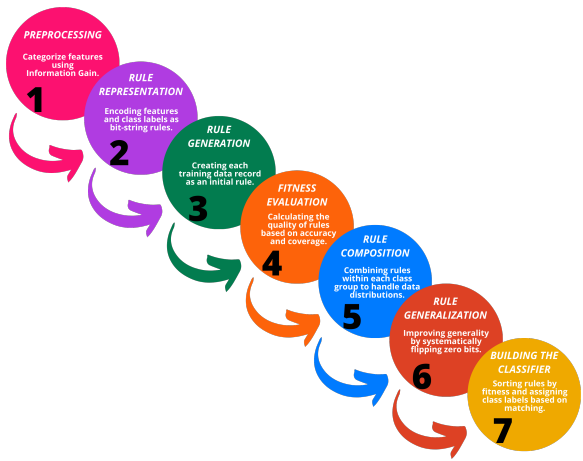

**Figure 1: Flow Diagram of the RACER Algorithm**

As illustrated in Figure 1, the algorithm is structured into several phases: preprocessing, rule representation, initial rule creation, fitness evaluation, rule composition, rule generalization, and classifier construction. The process begins by converting numerical data into categorical forms. RACER then represents rules using a bit string format, where each bit indicates the presence or absence of a specific attribute value, effectively encoding both features and class labels. This one-bit-per-value approach ensures uniform rule lengths, facilitating more efficient comparisons and calculations. For a dataset with $n$ features ($F_1, \ldots, F_n$) and $m$ class attributes, each rule has a fixed size of $(\sum_{i=1}^{n} n_i + m)$ bits, where $\sum_{i=1}^{n} n_i$ represents the total number of unique values across the features.

To illustrate, consider a record $R$ that reflects certain conditions, featuring $F_1$ to $F_5$ and a class label $F_C$, along with the following domains:

$$F_1 \in \{\text{morning, evening}\}$$
$$F_2 \in \{\text{clear, cloudy}\}$$
$$F_3 \in \{\text{excellent, poor}\}$$
$$F_4 \in \{\text{warm, cool}\}$$
$$F_5 \in \{\text{medium, high}\}$$
$$F_C \in \{\text{yes, no}\}$$

Here, $R$ will be represented in 12 bits, i.e., 2 bits for $F_1$, 2 for $F_2$, 2 for $F_3$, 2 for $F_4$, 2 for $F_5$, and 2 for the class label, which could be shown in the following format:

| $F_1$ | $F_2$ | $F_3$ | $F_4$ | $F_5$ | $F_C$ |
|---|---|---|---|---|---|
| 10 | 10 | 01 | 10 | 11 | 10 |

That is, if ($F_1$ = morning) and ($F_2$ = clear) and ($F_3$ = poor) and ($F_4$ = warm) and ($F_5$ = medium or high), then ($F_C$ = true).

In the next step, the algorithm generates initial rules from each training instance and evaluates their validity using a fitness function. These rules are then combined randomly using the logical "OR" operation. The algorithm further generalizes each rule by flipping all zero bits to one. When a bit flip leads to a rule with improved fitness, the algorithm updates the rule and continues with the remaining bits, focusing only on enhancing that specific rule.

The effectiveness of the algorithm heavily relies on the quality of these initial rules, which is influenced by the dataset's characteristics. Since the rules are based on the training data, RACER's search space is restricted to these initial configurations. This limitation is especially significant in high-dimensional, low-sample-size datasets, where the number of features exceeds the number of observations. Such conditions create challenges like multicollinearity and insufficient data for accurate parameter estimation, complicating analysis for many algorithms, including RACER.

This is where I propose RUMC.

## 4 THE PROPOSED CLASSIFIER: RUMC

The RUle Mutation Classifier (RUMC) aims to produce better initial rules using mutation techniques of evolutionary algorithms. This will be based on generating a modified set of initial rules, facilitating a more controlled process of rule generalization, and leading to more robust options for rule merging. The pseudocode for the RUMC classifier is illustrated in Algorithm 1.

**Algorithm 1** RUMC algorithm

1: **Input:** Training dataset $D_{\text{train}}$
2: **Output:** The set of decision rules $R_{\text{set}}$
3: **procedure** RUMC($D_{\text{train}}$)
4:     Preprocessing
5:     Create *Initial_Rules*
6:     *Extant_Rules* = *Initial_Rules*
7:     Rule Mutation
8:     Primary Generalization of *Extant_Rules*
9:     **for** $i = 1$ to $|Extant\_Rules|$ **do**
10:       **for** $j = i + 1$ to $|Extant\_Rules|$ **do**
11:         *Composed_Rule* = Rule Composition($R_i, R_{i+1}$)
12:         **if** *Composed_Rule* is better than $R_i$ and $R_{i+1}$ **then**
13:           Replace both $R_i$ and $R_{i+1}$ with *Composed_Rule*
14:           Omit all covered rules with *Composed_Rule*
15:           Update *Extant_Rules*
16:         **end if**
17:       **end for**
18:     **end for**
19:     Secondary Generalization of *Extant_Rules*
20:     Sort the *Extant_Rules* Based on Fitness
21:     $R_{\text{set}}$ = *Extant_Rules*
22: **end procedure**

The algorithm shares key similarities with the previously discussed RACER steps. After completing preprocessing, rule representation, and the initial rule creation, RUMC introduces a unique rule mutation phase. Following this, the algorithm continues with fitness evaluation and a new step for generalizing the primary rule. The subsequent phases, including rule composition, secondary rule generalization, and classifier construction, remain consistent with the original framework. The following sections will provide a detailed exploration of these processes and their technical intricacies.

### 4.1 Preprocessing

RUMC requires categorical domains for its features to operate effectively. To achieve this, I employ the Information Gain technique to convert continuous features into discrete categories [30]. The feature that yields the greatest reduction in uncertainty is chosen as the optimal split point. The initial step in applying this method for feature $F$ within the training dataset $D_{\text{train}}$, which contains $N$ records, involves sorting the values of $F$ in ascending order. According to the Information Gain criterion, the split point, denoted as MidPoint$_i$, is defined as the midpoint between two adjacent values $m_i$ and $m_{i+1}$ in $F$:

$$\text{MidPoint}_i = \frac{m_i + m_{i+1}}{2}, \quad \text{for} \quad i = 1, 2, \ldots, N \quad (1)$$

The function Info($D_{\text{train}}$), where $c$ is the count of classes and $P_i$ is the likelihood that an instance is in that particular class, is defined for each MidPoint$_i$ in the following way:

$$\text{Info}(D_{\text{train}}) = -\sum_{i=1}^{c} P_i \log_2(P_i) \quad (2)$$

Now, to calculate Info$_F(D_{\text{train}})$, where $D_1$ is a group of records with $F \leq$ MidPoint$_i$ and $D_2$ is a group of records with $F >$ MidPoint$_i$, the function Info$_F(D_{\text{train}})$ will be:

$$\text{Info}_F(D_{\text{train}}) = \frac{|D_1|}{|D_{\text{train}}|} \cdot \text{Info}(D_1) + \frac{|D_2|}{|D_{\text{train}}|} \cdot \text{Info}(D_2) \quad (3)$$

So, for each MidPoint$_i$ in $F$, we compute Info$_F(D_{\text{train}})$, and the one with the lowest Info$_F(D_{\text{train}})$ will act as the optimal split point.

### 4.2 Rule Representation

RUMC employs a bit-wise representation of rules, similar to the method used by RACER. In this framework, each feature is represented by bits corresponding to the number of its possible values. A bit set to one indicates that the feature value is true, while a zero signifies its false.

### 4.3 Initial Rule Generation

To create initial rules, the algorithm encodes each record in the dataset as a binary string, representing specific values with sequences of 0s and 1s. For example, with a training dataset containing $N$ records, the algorithm generates $N$ individual initial rules, collectively forming the rule set.

### 4.4 Rule Mutation

RUMC employs a specialized mutation phase to enhance the quality and diversity of its initial rules. This approach identifies false bits in each rule and iteratively flips them to true while checking for fitness improvements. If a modification results in better fitness, the rule and its associated fitness score are updated accordingly. This strategy allows the algorithm to replace underperforming rules with mutated alternatives, balancing generality and effectiveness.

The mutation function operates by identifying the indices of false bits in each rule. It creates copies of the original rule, flipping one false bit in each copy at a time, and evaluates their fitness. If any modified rule shows higher fitness than the original, it replaces the latter in the rule set. This process continues until all false bits have been examined. If no improvements are found or a class conflict arises, the original rule is retained. For instance, in a rule with four false indices, the mutation function generates four copies, each with one false bit flipped, and assesses their fitness. The rule with the highest fitness is adopted, ensuring that all potential enhancements are explored while maintaining the integrity of the rules.

Generating enhanced rules broadens the scope for rule merging, facilitating a wider range of scenarios within the search space and increasing the potential for an optimized final rule set. Also, zero bits are less represented in the training data, indicating greater flexibility in bit positions. Flipping these bits broadens the matching range, thereby improving rule performance and adaptability to new, unseen data. The implementation of the rule mutation phase in here is detailed in Algorithm 2.

### 4.5 Fitness Evaluation

The quality of each rule is evaluated using a fitness function that considers both accuracy and coverage. This function is defined as follows:

$$Fitness(R_i) = \alpha \times Accuracy(R_i) + \beta \times Coverage(R_i) \quad (4)$$

**Algorithm 2** Rule Mutation Phase of RUMC

1: **Input:** Extant Rules
2: **Output:** Updated list of Extant Rules
3: **procedure** RUM(Extant Rules)
4:    **for** each rule $i$ in Extant Rules **do**
5:       Identify all 0-valued bit indices in rule $i$
6:       **for** each 0-bit index $j$ **do**
7:          Copy rule $i$ to Temporary Rule
8:          Flip bit at index $j$ in Temporary Rule
9:          Calculate Fitness of Temporary Rule
10:          **if** Fitness improves **then**
11:             Replace Current Rule with Temporary Rule
12:             Update Fitness of Current Rule
13:          **end if**
14:       **end for**
15:    **end for**
16: **end procedure**

In this context, $R_i$ denotes a rule within the rule set, while $\alpha$ and $\beta$ serve as weighting factors that reflect the importance of Accuracy and Coverage, respectively. These weights must adhere to the conditions $\alpha, \beta > 0$ and $\alpha + \beta = 1$.

Accuracy($R_i$) measures the proportion of correctly classified records among those covered by the rule $R_i$. It is computed as the ratio of correctly classified records ($n_{\text{correct}}$) to the total number covered ($n_{\text{covers}}$) by the $R_i$ rule:

$$\text{Accuracy}(R_i) = \frac{n_{\text{correct}}}{n_{\text{covers}}} \quad (5)$$

On the other hand, Coverage($R_i$) measures the proportion of records in the training dataset $D_{\text{train}}$ that rule $R_i$ covers. It is calculated as the ratio of the number of records the rule $R_i$ covers ($n_{\text{covers}}$) to the total number of records in the training dataset ($|D_{\text{train}}|$).

$$\text{Coverage}(R_i) = \frac{n_{\text{covers}}}{|D_{\text{train}}|} \quad (6)$$

This method ensures that the resultant rule sets yield precise predictions while remaining applicable to extensive datasets.

### 4.6 Primary Rule Generalization

Akin to generalization in evolutionary algorithms, RUMC uses a two-step method to improve coverage. In the first phase, it assesses both existing and newly mutated rules. The process starts by locating the zero bits in rule $R$. Changing these bits to one can improve the generality of $R$. It modifies the first zero bit, and if this change leads to a better fitness value, the update is accepted, and the next zero bit is analyzed. If there's no improvement, the change is ignored, and the next bit is considered. This process continues for all zero bits in $R$. After finishing with the current rule, RUMC moves on to the next rule and repeats the steps until all bits in all rules have been processed.

The key difference between this phase and the mutation phase lies in how the zero bits are handled. In the mutation phase, all zero bits are flipped simultaneously, selecting the best-performing modification. Conversely, the generalization phase focuses on one bit at a time, enabling a more systematic approach for generality.

### 4.7 Rule Composition

The initial rules are effective for the training set, but their specificity makes them less effective on the test set. To improve this, RUMC adopts a strategy that groups the initial rules by class labels and combines them within each group. This approach seeks to create broader rules that can handle a larger variety of data. Here's an example to illustrate this method:

| $F_1$ | $F_2$ | $F_3$ | $F_4$ | $F_5$ | $F_C$ |
|---|---|---|---|---|---|
| 10 | 10 | 01 | 10 | 11 | 10 |
| 01 | 10 | 01 | 11 | 01 | 10 |

After combining these two rules, a new rule will be generated based on the logical "OR" operation of those individual rules, which is:

| $F_1$ | $F_2$ | $F_3$ | $F_4$ | $F_5$ | $F_C$ |
|---|---|---|---|---|---|
| 11 | 10 | 01 | 11 | 11 | 10 |

This rule can be simplified further. By disregarding the non-informative features $F_1$, $F_4$, and $F_5$, which consist solely of 1s, merging the two original 12-bit rules will yield a single 6-bit rule, as illustrated below:

| $F_2$ | $F_3$ | $F_C$ |
|---|---|---|
| 10 | 01 | 10 |

The process begins with the first rule in each group, which is randomly merged with other rules. After each merge, the fitness of the resulting rule is evaluated. If it surpasses that of the original rules, it replaces them. This cycle repeats until no further improvements can be made.

### 4.8 Secondary Rule Generalization

Since the previous phases, we have developed a new set of rules that requires further generalization. We will employ a method similar to the initial generalization phase to achieve this goal once again.

### 4.9 Building the Classifier

In building the classifier, the first step is to sort the rules by their fitness values in descending order, prioritizing those generated earlier in case of ties. When classifying new records, the algorithm checks each instance against the highest-fitness rule first. If the instance meets the rule's conditions, it receives the corresponding class label. This continues until a matching rule is found.

Regarding complexity, the rule mutation phase of RUMC generates multiple copies of the arrays and evaluates their fitness, resulting in higher time complexity compared to the basic RACER. Additionally, its space complexity increases due to the memory needed for these copies. However, this is balanced by improved exploration of the search space and the potential for greater accuracy.

## 5 EXPERIMENTS

The performance of RUMC algorithm was analyzed, using a microcomputer with an Intel® Core™ i5-3337U CPU at 1.80GHz and 8 GB of RAM. The study involved forty datasets, employing ten-fold cross-validation with a seed value of one to assess mean classification accuracy. Other constants used in the code is also presented in Table 1. The comparison was conducted against twenty other classifiers, ranging from old to new approaches, using WEKA Workbench 3.9.6 [17] for all except ROPAC-L, ROPAC-M, and RACER which were implemented in Python 3.13. The classifiers evaluated included ROPAC-L, ROPAC-M, ForestPA, MLP, LMT, Random Forest, Optimized Forest, RACER, SPAARC, PART, Bagging, C4.5, JRip, CHAID*, ForEx++, CART, SMO, IBk, Naïve Bayes, and CSForest.

Table 1: Constants and Their Values

| Constant | Value |
| --- | --- |
| Accuracy Weight ($\alpha$) | 0.99 |
| Coverage Weight ($\beta$) | 0.01 |
| Coverage Weight for Rule Composition ($\gamma$) | 0.6 |

## 5.1 Data

The evaluation utilized 40 datasets from the UCI Machine Learning Repository [2] and OpenML [41]. These datasets were selected for their diversity and relevance. Table 2 summarizes key details, including the number of instances, missing values, feature types, and target classes.

Table 2: Dataset Characteristics

| # | Dataset Name | # of inst. | # of feat. | # of classes | # of mis. val. | # of num. attr. | # of cat. attr. |
| --- | --- | --- | --- | --- | --- | --- | --- |
| 1 | chscase_vine1 | 52 | 10 | 2 | 0 | 9 | 1 |
| 2 | dbworld-bodies | 64 | 4703 | 2 | 0 | 0 | 4703 |
| 3 | pyrim | 74 | 28 | 2 | 0 | 27 | 1 |
| 4 | kidney | 76 | 7 | 2 | 0 | 3 | 4 |
| 5 | analcatdata_asbestos | 83 | 4 | 2 | 0 | 1 | 3 |
| 6 | baskball | 96 | 5 | 2 | 0 | 4 | 1 |
| 7 | analcatdata_chlamydia | 100 | 4 | 2 | 0 | 0 | 4 |
| 8 | fertility | 100 | 10 | 2 | 0 | 9 | 1 |
| 9 | molecular-biology_promoters | 106 | 58 | 2 | 0 | 0 | 58 |
| 10 | fruitfly | 125 | 5 | 2 | 0 | 2 | 3 |
| 11 | mux6 | 128 | 7 | 2 | 0 | 0 | 7 |
| 12 | analcatdata_wildcat | 163 | 6 | 2 | 0 | 3 | 3 |
| 13 | servo | 167 | 5 | 2 | 0 | 0 | 5 |
| 14 | parkinsons | 195 | 23 | 2 | 0 | 22 | 1 |
| 15 | pwLinear | 200 | 11 | 2 | 0 | 10 | 1 |
| 16 | cpu | 209 | 8 | 2 | 0 | 6 | 2 |
| 17 | seeds | 210 | 8 | 3 | 0 | 7 | 1 |
| 18 | chatfield_4 | 235 | 13 | 2 | 0 | 12 | 1 |
| 19 | heart-statlog | 270 | 14 | 2 | 0 | 13 | 1 |
| 20 | breastTumor | 286 | 10 | 2 | 9 | 1 | 9 |
| 21 | heart-h | 294 | 14 | 2 | 782 | 6 | 8 |
| 22 | cleveland | 303 | 14 | 2 | 6 | 6 | 8 |
| 23 | ecoli | 327 | 8 | 5 | 0 | 7 | 1 |
| 24 | liver-disorders | 345 | 7 | 2 | 0 | 6 | 0 |
| 25 | vote | 435 | 17 | 2 | 392 | 0 | 17 |
| 26 | threeOf9 | 512 | 10 | 2 | 0 | 0 | 10 |
| 27 | kc2 | 522 | 22 | 2 | 0 | 21 | 1 |
| 28 | wdbc (Breast Cancer Wisconsin) | 569 | 31 | 2 | 0 | 30 | 1 |
| 29 | monks-problems-2 | 601 | 7 | 2 | 0 | 0 | 7 |
| 30 | eucalyptus | 736 | 20 | 2 | 448 | 14 | 6 |
| 31 | diabetes | 768 | 9 | 2 | 0 | 8 | 1 |
| 32 | stock | 950 | 10 | 2 | 0 | 9 | 1 |
| 33 | tokyo1 | 959 | 45 | 2 | 0 | 42 | 3 |
| 34 | xd6 | 973 | 10 | 2 | 0 | 0 | 10 |
| 35 | flare | 1066 | 11 | 2 | 0 | 0 | 11 |
| 36 | parity5_plus_5 | 1124 | 11 | 2 | 0 | 0 | 11 |
| 37 | cmc | 1473 | 10 | 2 | 0 | 2 | 8 |
| 38 | kr-vs-kp | 3196 | 37 | 2 | 0 | 0 | 37 |
| 39 | led7 | 3200 | 8 | 10 | 0 | 0 | 8 |
| 40 | nursery | 12960 | 9 | 5 | 0 | 0 | 9 |

I select datasets with feature counts ranging from 4 to 4703 for my analysis. This collection includes datasets with categorical features, varying from 0 to 4703, and numerical features, ranging from 0 to 42. Notably, five datasets have missing values, with counts between 6 and 782. The datasets also exhibit diverse target labels, with class counts ranging from 2 to 10. In terms of size, the collection features both compact datasets, such as "chscase_vine1" with 52 records, and larger datasets like "nursery," with 12,960 records.

## 5.2 Evaluation Criteria

The accuracy metric, which gauges how well the model assigns class labels to instances, was used to evaluate the efficacy of the RUMC algorithm. The following formula is used to determine accuracy:

$$\text{Accuracy} = \frac{\text{Number of correctly classified instances}}{\text{Total number of instances}} \quad (7)$$

## 5.3 Results

I evaluated the performance of RUMC against other popular classification techniques, focusing on mean accuracy. Table 3 summarizes these results, highlighting the best accuracy in bold for easy identification, which include RUMC and twenty other classifiers. The classifiers span various approaches, including Naïve Bayes, SMO (SVM), IBk (KNN), MLP (neural networks), ensemble methods (Bagging, Random Forest, Optimized Forest, CSForest), rule-based methods (JRip, PART, RACER), and decision trees (C4.5, CART, CHAID*, ForestPA, LMT, SPAARC, ForEx++). The best accuracy for each row is highlighted in bold.

According to Table 3, RUMC surpasses all other algorithms in average accuracy, achieving an improvement of 0.1% over ROPAC-L and 10.65% over CSForest. This trend continues with rule-based classifiers; while JRip reaches an accuracy of 80.49% and RACER achieves 81.6%, RUMC demonstrate impressive rates of 84.33%. Algorithms like Naïve Bayes, IBk, PART, Bagging, CSForest, and JRip consistently underperform. Notably, Naïve Bayes and IBk show low accuracy in most datasets. Decision tree algorithms exhibit considerable variability; traditional methods like C4.5 and CART are often outperformed by contemporary techniques like ForestPA, suggesting that while conventional decision trees may be effective for general applications, ensemble methods may excel in specific scenarios [9].

When comparing the overall performance of algorithms, notable patterns emerge: neural networks (MLP) often rank highly, while lazy learners (IBk from KNN) typically score lower. It's also worth mentioning that ROPAC-L performs well, although RUMC stands out even more. However, no single algorithm consistently outperforms others, as performance varies by dataset [13, 44]. Generally, based on my experiments, RUMC proves to be a reliable choice for a wide array of classification tasks.

The exceptional performance of RUMC can be attributed to its unique rule mutation process. This algorithm stands out by expanding the search space and employing an iterative approach to refine rules with mutation inspired by genetic evolution. While traditional classifiers focus on individual features or adopt a broad perspective, they often neglect the critical analysis and generalization of initial rules. RUMC's consistent performance across diverse datasets highlights its adaptability, making it a strong candidate for various real-world applications.

# Table 3: Accuracy Comparison of Classifiers

| # | Dataset | RUMC | ROPAC-L | ROPAC-M | ForestPA | MLP | LMT | Random Forest | Optimized Forest | RACER | SPAARC | PART | Bagging | C4.5 | JRip | CHAID* | ForEx++ | CART | SMO | IBk | Naïve Bayes | CSForest |
|---|---------|------|---------|---------|----------|-----|-----|---------------|------------------|-------|--------|------|---------|------|------|--------|---------|------|-----|-----|-------------|----------|
| 1 | chscase_vine1 | 83.67 | 83.67 | 83.67 | 84.62 | **84.62** | 76.92 | 82.69 | 82.69 | 79.67 | 76.92 | 76.92 | 82.69 | 73.08 | 80.77 | 73.08 | 84.62 | 75 | **84.62** | 76.92 | 82.69 | 55.77 |
| 2 | dbworld-bodies | **87.62** | **87.62** | **87.62** | 85.94 | 81.77 | 81.25 | 82.81 | 82.81 | **87.62** | 84.38 | 81.25 | 82.81 | 79.69 | 84.38 | 68.75 | 85.94 | 79.69 | 87.5 | 59.38 | 75 | 71.88 |
| 3 | pyrim | 89.11 | 89.11 | 89.11 | 83.78 | 90.54 | **94.59** | 87.84 | 87.84 | 86.43 | 81.08 | 87.84 | 81.08 | 82.43 | 83.78 | 87.84 | 89.19 | 86.49 | 90.54 | 87.84 | 86.49 | 72.97 |
| 4 | kidney | 77.68 | 75 | 73.75 | 72.37 | 71.05 | 65.79 | 76.32 | 75 | 71.25 | 72.37 | 69.74 | 75 | 68.42 | 67.11 | 63.16 | **78.95** | 60.53 | 53.95 | 71.05 | 69.74 | 56.58 |
| 5 | analcatdata_asbestos | **76.11** | **76.11** | **76.11** | 75.90 | 72.29 | 73.49 | 71.08 | 69.88 | 72.5 | 69.88 | 73.49 | 68.67 | 72.29 | 71.08 | 72.29 | 75.90 | 72.29 | 72.29 | 69.88 | 73.49 | 75.90 |
| 6 | baskball | 74.22 | 74.22 | 74.22 | 71.88 | 69.79 | **76.04** | 67.71 | 67.71 | 59.56 | 67.71 | 68.75 | 64.58 | 69.79 | 67.71 | 69.79 | 69.79 | 67.71 | 70.83 | 60.42 | 71.88 | 48.96 |
| 7 | analcatdata_chlamydia | 90 | 90 | 84 | 87 | 86 | **91** | 89 | 87 | 83 | 88 | 80 | 75 | 81 | 74 | 86 | 81 | 77 | 85 | 82 | 87 | 67 |
| 8 | fertility | 86 | 86 | 86 | 88 | **90** | 89 | 86 | 86 | 85 | 88 | 88 | 88 | 85 | 87 | 88 | 88 | 84 | 88 | 83 | 88 | 78 |
| 9 | molecular-biology_promoters | 68 | 68 | 68 | 67.92 | 64.15 | 60.38 | 67.92 | 67.92 | 68 | 67.92 | 60.38 | 67.92 | 49.06 | 62.26 | 67.92 | 67.92 | 67.92 | 62.26 | 66.98 | **69.81** | 66.98 |
| 10 | fruitfly | **64.29** | **64.29** | **64.29** | 59.2 | 48.8 | 60 | 49.6 | 51.2 | 64.23 | 60 | 53.6 | 56 | 61.6 | 58.4 | 60.8 | 60 | 61.6 | 60.8 | 48.8 | 58.4 | 39.2 |
| 11 | mux6 | **100** | **100** | **100** | 98.44 | 97.66 | 90.63 | **100** | **100** | **100** | 78.13 | 98.44 | 89.84 | 90.63 | **100** | 84.38 | 67.19 | 82.03 | 67.97 | 98.44 | 57.81 | 59.38 |
| 12 | analcatdata_wildcat | 76.69 | 76.69 | 76.69 | 77.91 | **78.53** | 75.46 | 76.07 | 76.07 | 71.07 | 77.91 | 76.69 | 74.23 | 74.23 | 74.23 | 71.78 | 73.01 | 71.78 | 72.39 | 73.01 | 74.85 | 74.23 |
| 13 | servo | 92.83 | 92.83 | 92.87 | **94.61** | 93.41 | 91.62 | 92.81 | 92.81 | 91.58 | 93.41 | 94.01 | 91.02 | 91.62 | 92.81 | 92.81 | 92.81 | 89.82 | 93.41 | 92.22 | 92.81 | 92.81 |
| 14 | parkinsons | 87.32 | 87.32 | 87.34 | 88.72 | 90.77 | 86.15 | 92.82 | 92.31 | 86.29 | 86.67 | 81.03 | 88.72 | 80.51 | 87.69 | 86.15 | 84.62 | 85.64 | 87.18 | **96.41** | 69.23 | 84.62 |
| 15 | pwLinear | 86.5 | 85 | 87.5 | 89.5 | 88.5 | 87.5 | **91** | **91** | 83.5 | 88 | 88 | 85.5 | 87 | 87 | 85.5 | 85 | 81 | 85.5 | 83.5 | 71.5 | 78.5 |
| 16 | cpu | 97.62 | 97.62 | 98.1 | 96.17 | 97.13 | **99.52** | 96.17 | 96.65 | 97.14 | 95.69 | 95.69 | 96.17 | 97.61 | 95.22 | 94.26 | 97.61 | 93.3 | 95.22 | 94.74 | 95.22 | 96.65 |
| 17 | seeds | 91.43 | 91.43 | 90.48 | 89.52 | **95.24** | **95.24** | 94.29 | 94.29 | 88.1 | 90.95 | 92.86 | 92.86 | 91.9 | 90.48 | 88.57 | 88.10 | 90 | 93.81 | 94.29 | 91.43 | 91.90 |
| 18 | chatfield_4 | 89.33 | 89.76 | 89.33 | 88.09 | 88.94 | 87.66 | 87.66 | 88.51 | 88.04 | **90.21** | 84.26 | 88.51 | 89.36 | 85.96 | **90.21** | 89.79 | 88.09 | 86.38 | 83.4 | 87.66 | 85.96 |
| 19 | heart-statlog | **84.44** | 82.59 | 82.59 | 82.59 | 78.15 | 83.33 | 81.48 | 80.74 | 79.26 | 78.89 | 73.33 | 79.26 | 76.67 | 78.89 | 76.30 | 83.70 | 76.67 | 84.07 | 75.19 | 83.7 | 71.11 |
| 20 | breastTumor | 55.94 | 57.34 | 58.42 | 55.24 | 45.8 | **63.64** | 50.7 | 50.35 | 53.53 | 56.99 | 52.1 | 53.5 | 54.9 | 55.94 | 56.64 | 55.94 | 57.69 | **63.64** | 47.55 | 60.49 | 57.69 |
| 21 | heart-h | 84.67 | 83.67 | 84.34 | 82.65 | **85.03** | **85.03** | 81.97 | 81.97 | 81.68 | 77.55 | 80.95 | 79.25 | 80.95 | 78.91 | 78.23 | 39.12 | 77.55 | 82.65 | 76.87 | 83.67 | 39.12 |
| 22 | cleveland | 79.19 | 79.88 | 79.23 | 81.52 | 77.56 | 82.51 | 81.85 | 82.84 | 79.55 | 78.22 | 78.55 | 81.52 | 76.57 | 76.57 | 80.20 | 82.51 | 77.56 | **83.17** | 75.58 | 82.51 | 67.66 |
| 23 | ecoli | 82.25 | 81.95 | 82.25 | 88.07 | 86.85 | **88.99** | 86.54 | 87.16 | 73.99 | 82.26 | 84.4 | 85.32 | 83.49 | 85.93 | 83.18 | 84.10 | 82.57 | 84.1 | 82.26 | 87.46 | 85.63 |
| 24 | liver-disorders | 68.41 | 68.41 | 68.7 | 69.86 | 71.59 | 66.38 | 73.04 | **74.20** | 59.13 | 66.38 | 63.77 | 69.57 | 68.7 | 64.64 | 65.22 | 66.96 | 64.06 | 58.26 | 62.9 | 55.36 | 59.13 |
| 25 | vote | 95.14 | 95.15 | 95.15 | 95.63 | 94.71 | **96.78** | 96.09 | 96.09 | 94.48 | 95.40 | 94.71 | 95.63 | 96.32 | 95.4 | 94.02 | 95.63 | 95.4 | 96.09 | 92.41 | 90.11 | 95.17 |
| 26 | threeOf9 | **100** | **100** | 99.8 | 99.61 | 97.66 | 98.83 | 99.61 | 99.61 | 99.61 | 95.90 | 99.02 | 99.02 | 97.85 | **100** | 96.88 | 98.83 | 95.9 | 80.08 | 99.8 | 80.66 | 78.52 |
| 27 | kc2 | 83.94 | 83.74 | 84.13 | 83.52 | **84.67** | 84.29 | 83.33 | 83.14 | 83.94 | 82.95 | 82.18 | 83.72 | 81.42 | 82.18 | 81.99 | 84.48 | 81.61 | 82.76 | 80.46 | 83.52 | 79.12 |
| 28 | wdbc (Breast Cancer Wisconsin) | 96.48 | 96.83 | 96.48 | 94.55 | 96.13 | 97.19 | 95.96 | 95.96 | 95.25 | 93.50 | 93.5 | 95.25 | 93.32 | 92.62 | 93.85 | 95.96 | 92.27 | **97.72** | 95.96 | 92.97 | 93.32 |
| 29 | monks-problems-2 | 73.22 | 72.73 | 73.73 | 68.72 | **77.54** | 76.54 | 44.43 | 45.92 | 67.22 | 74.71 | 61.56 | 48.42 | 62.23 | 57.07 | 49.92 | 54.74 | 49.92 | 43.26 | 50.25 | 45.59 | 50.25 |
| 30 | eucalyptus | 75.68 | 75.82 | 75.55 | **76.36** | 73.64 | 75.95 | 74.86 | 75 | 75.54 | 75.41 | 75 | 71.2 | 74.46 | 72.01 | 69.43 | 63.86 | 70.38 | 75.27 | 69.84 | 65.63 | 70.92 |
| 31 | diabetes | 72.13 | 72.4 | 70.05 | 73.05 | 75.39 | **77.47** | 75.78 | 75.39 | 66.93 | 74.48 | 75.26 | 75.78 | 73.83 | 76.04 | 73.83 | 74.74 | 75.26 | 77.34 | 70.18 | 76.3 | 67.45 |
| 32 | stock | 92.84 | 92.84 | 92.84 | 95.47 | 94.84 | 95.89 | 96.74 | **96.84** | 88.32 | 94.84 | 95.37 | 95.37 | 95.79 | 95.37 | 94.32 | 95.47 | 94.21 | 84.21 | 96.32 | 70.84 | 92.63 |
| 33 | tokyo1 | **93.01** | 92.7 | 92.7 | 92.60 | 91.66 | 92.28 | 92.7 | 92.81 | 92.91 | 91.76 | 90.51 | 92.28 | 91.35 | 90.93 | 92.49 | 92.49 | 92.28 | 91.87 | 91.35 | 90.62 | 92.91 |
| 34 | xd6 | **100** | **100** | **100** | 100 | 98.77 | 99.9 | **100** | **100** | 99.9 | 99.90 | **100** | **100** | 99.9 | **100** | 99.38 | **100** | 98.87 | 78.83 | **100** | 81.4 | 94.55 |
| 35 | flare | 81.7 | 81.51 | 81.7 | 82.08 | 79.74 | 82.55 | 80.86 | 81.14 | 82.17 | 82.93 | 81.33 | 82.55 | 82.08 | 82.46 | 82.74 | 82.55 | 82.74 | **83.68** | 80.58 | 80.02 | 75.05 |
| 36 | parity5_plus_5 | **100** | **100** | **100** | 80.52 | 94.66 | 53.02 | 69.75 | 67.97 | **100** | 59.79 | 91.01 | 73.31 | 92.62 | 48.49 | 50.44 | 38.08 | 58.19 | 46 | 59.34 | 40.21 | 51.69 |
| 37 | cmc | 65.58 | 66.73 | 66.67 | 69.59 | 67.35 | 69.31 | 67.82 | 66.60 | 65.17 | 70.88 | 66.46 | 71.15 | 68.97 | 70.88 | **71.22** | 70.20 | 69.72 | 67.14 | 59.67 | 65.78 | 66.19 |
| 38 | kr-vs-kp | 99.03 | 98.78 | 98.97 | 99.09 | 99.34 | **99.75** | 99.09 | 99.09 | 99.22 | 99.31 | 99.06 | 99.06 | 99.44 | 99.19 | 99.06 | 97.68 | 99 | 95.43 | 96.28 | 87.89 | 97.37 |
| 39 | led7 | 71.34 | 71.59 | 71.44 | **73.94** | 73.56 | 73.66 | 73.13 | 73.28 | 63.5 | 73.16 | 73.56 | 72.88 | 73.34 | 69.16 | 72.88 | 73.69 | 72.97 | 73.56 | 73.31 | 73.16 | 73.47 |
| 40 | nursery | **99.75** | 99.71 | **99.75** | 99.63 | 99.73 | 98.99 | 99.07 | 99.05 | 99.63 | 99.58 | 99.21 | 97.34 | 97.05 | 96.84 | 97.61 | 95.25 | 95.96 | 93.08 | 98.38 | 90.32 | 90.86 |
| | AVERAGE | 84.33 | 84.23 | 84.09 | 83.6 | 83.34 | 83.11 | 82.41 | 82.37 | 81.6 | 81.55 | 81.54 | 81.25 | 81.16 | 80.49 | 79.78 | 79.64 | 79.37 | 79 | 78.92 | 76.78 | 73.68 |

Melvin Mokhtari

# 6 CONCLUSION

The RUle Mutation Classifier (RUMC) is a novel rule-based classification algorithm, utilizing rule mutation techniques inspired by evolutionary algorithms, alongside a two-tier generalization strategy. This study conducts a comprehensive evaluation of RUMC across forty diverse datasets, aiming to showcase its advantages over prior algorithms and its performance in various scenarios, including those with high dimensionality and limited sample sizes. The findings reveal that RUMC's enhancements allow it to outperform a wide range of established algorithms in terms of average classification accuracy.

Future research could explore various pathways for improvement. One potential direction involves incorporating a dynamic fitness function that adjusts during iterative training cycles. Exploring other discretization methods beyond Information Gain may also uncover new possibilities.

## DATA AVAILABILITY

The code and data supporting this study are available under the MIT License and Attribution (CC BY) and can be accessed at https://github.com/MelvinMo/RUMC-RUle-Mutation-Classifier.

## REFERENCES

[1] 1998. Generating accurate rule sets without global optimization. *University of Waikato, Department of Computer Science* 144 (1 1998). https://hdl.handle.net/10289/1047
[2] 2017. UCI Machine Learning Repository (Technical Report, University of California, Irvine, School of Information and Computer Sciences). http://archive.ics.uci.edu/ml
[3] Md Nasim Adnan and Md Zahidul Islam. 2016. Optimizing the number of trees in a decision forest to discover a subforest with high ensemble accuracy using a genetic algorithm. *Knowledge-Based Systems* 110 (7 2016), 86–97. https://doi.org/10.1016/j.knosys.2016.07.016
[4] Md Nasim Adnan and Md Zahidul Islam. 2017. Forest PA : Constructing a decision forest by penalizing attributes used in previous trees. *Expert Systems with Applications* 89 (8 2017), 389–403. https://doi.org/10.1016/j.eswa.2017.08.002
[5] Md Nasim Adnan and Md Zahidul Islam. 2017. ForEx++: A New Framework for Knowledge Discovery from Decision Forests. *AJIS. Australasian journal of information systems/AJIS. Australian journal of information systems/Australian journal of information systems* 21 (11 2017). https://doi.org/10.3127/ajis.v21i0.1539
[6] Hayder Naser Khraibet Al-Behadili, Ku Ruhana Ku-Mahamud, and Rafid Sagban. 2019. Annealing strategy for an enhance rule pruning technique in ACO-Based rule classification. *Indonesian Journal of Electrical Engineering and Computer Science* 16, 3 (12 2019), 1499. https://doi.org/10.11591/ijeecs.v16.i3.pp1499-1507
[7] Javad Basiri, Fattaneh Taghiyareh, and Heshaam Faili. 2019. RACER: accurate and efficient classification based on rule aggregation approach. *Neural Computing and Applications* 31, 3 (7 2019), 895–908. https://doi.org/10.1007/s00521-017-3117-2
[8] Katelyn Battista, Liqun Diao, Karen A. Patte, Joel A. Dubin, and Scott T. Leatherdale. 2023. Examining the use of decision trees in population health surveillance research: an application to youth mental health survey data in the COMPASS study. *Health Promotion and Chronic Disease Prevention in Canada* 43, 2 (2 2023), 73–86. https://doi.org/10.24095/hpcdp.43.2.03
[9] Jaison Bennet, Chilambuchelvan Arul Ganaprakasam, and Kannan Arputharaj. 2014. A discrete Wavelet based feature extraction and hybrid classification technique for microarray data analysis. *The Scientific World JOURNAL* 2014 (1 2014), 1–9. https://doi.org/10.1155/2014/195470
[10] Rajarshi Bhattacharjee and Naresh Manwani. 2020. *Online algorithms for multiclass classification using partial labels.* 249–260 pages. https://doi.org/10.1007/978-3-030-47426-3_20
[11] Jadzia Cendrowska. 1987. PRISM: An algorithm for inducing modular rules. *International Journal of Man-Machine Studies* 27, 4 (10 1987), 349–370. https://doi.org/10.1016/s0020-7373(87)80003-2
[12] William W. Cohen. 1995. *Fast effective rule induction.* 115–123 pages. https://doi.org/10.1016/b978-1-55860-377-6.50023-2
[13] Antoine Decoux, Loic Duron, Paul Habert, Victoire Roblot, Emina Arsovic, Guillaume Chassagnon, Armelle Arnoux, and Laure Fournier. 2023. Comparative performances of machine learning algorithms in radiomics and impacting factors. *Research Square (Research Square)* (4 2023). https://doi.org/10.21203/rs.3.rs-2677455/v1
[14] Shih-Lun Fang, Yuan-Kai Tu, Le Kang, Han-Wei Chen, Ting-Jung Chang, Min-Hwi Yao, and Bo-Jein Kuo. 2023. CART model to classify the drought status of diverse tomato genotypes by VPD, air temperature, and leaf–air temperature difference. *Scientific Reports* 13, 1 (1 2023). https://doi.org/10.1038/s41598-023-27798-8
[15] Takashi Fuchino, Takashi Harada, Ken Tanaka, and Kenji Mikawa. 2023. Computational complexity of Allow rule ordering and its greedy algorithm. *IEICE Transactions on Fundamentals of Electronics Communications and Computer Sciences* E106.A, 9 (3 2023), 1111–1118. https://doi.org/10.1587/transfun.2022dmp0006
[16] Johannes Fürnkranz and Gerhard Widmer. 1994. *Incremental reduced error pruning.* 70–77 pages. https://doi.org/10.1016/b978-1-55860-335-6.50017-9
[17] Mark Hall, Eibe Frank, Geoffrey Holmes, Bernhard Pfahringer, Peter Reutemann, and Ian H. Witten. 2009. The WEKA data mining software. *ACM SIGKDD Explorations Newsletter* 11, 1 (11 2009), 10–18. https://doi.org/10.1145/1656274.1656278
[18] Harshad Hegde, Ingrid Glurich, Aloksagar Panny, Jayanth G. Vedre, Jeffrey J. VanWormer, Richard Berg, Frank A. Scannapieco, Jeffrey Miecznikowski, and Amit Acharya. 2022. Identifying Pneumonia Subtypes from Electronic Health Records Using Rule-Based Algorithms. *Methods of Information in Medicine* 61, 01/02 (3 2022), 029–037. https://doi.org/10.1055/a-1801-2718
[19] Piotr Hońko. 2019. Binary classification rule generation from decomposed data. *International Journal of Intelligent Systems* 34, 12 (9 2019), 3123–3138. https://doi.org/10.1002/int.22181
[20] Igor Ibarguren, Aritz Lasarguren, Jesús M. Pérez, Javier Muguerza, Ibai Gurrutxaga, and Olatz Arbelaitz. 2016. BFPART: Best-First PART. *Information Sciences* 367-368 (7 2016), 927–952. https://doi.org/10.1016/j.ins.2016.07.023
[21] Ayad Jabba. 2021. Rule Induction with Iterated Local Search. *International journal of intelligent engineering and systems* 14, 4 (6 2021), 289–298. https://doi.org/10.22266/ijies2021.0831.26
[22] G. Kesavaraj and S. Sukumaran. 2013. A study on classification techniques in data mining. *2013 Fourth International Conference on Computing, Communications and Networking Technologies (ICCCNT)* (7 2013). https://doi.org/10.1109/icccnt.2013.6726842
[23] Dániel Koren, Laura Lőrincz, Sándor Kovács, Gabriella Kun-Farkas, Beáta Vecseriné Hegyes, and László Sipos. 2020. Comparison of supervised learning statistical methods for classifying commercial beers and identifying patterns. *Journal of Chemometrics* 34, 4 (1 2020). https://doi.org/10.1002/cem.3216
[24] Thien Le, Frederic Stahl, João Bártolo Gomes, Mohamed Medhat Gaber, and Giuseppe Di Fatta. 2014. *Computationally efficient Rule-Based classification for continuous streaming data.* 21–34 pages. https://doi.org/10.1007/978-3-319-12069-0_2
[25] Han Liu, Alexander Gegov, and Mihaela Cocea. 2016. *Complexity control in rule based models for classification in machine learning context.* 125–143 pages. https://doi.org/10.1007/978-3-319-46562-3_9
[26] Jana Lukáčová and Lenka Maličká. 2022. *Real estate tax revenue and its place in local government tax revenue.* https://doi.org/10.33542/spf22-0145-2-14
[27] Li Mao, Qidong Chen, and Jun Sun. 2020. Construction and Optimization of Fuzzy Rule-Based Classifier with a Swarm Intelligent Algorithm. *Mathematical Problems in Engineering* 2020 (4 2020), 1–12. https://doi.org/10.1155/2020/9319364
[28] Melvin Mokhtari and Alireza Basiri. 2024. ROPAC: Rule OPtimized Aggregation Classifier. *Expert Systems with Applications* 250 (4 2024), 123897. https://doi.org/10.1016/j.eswa.2024.123897
[29] Maryam M Najafabadi, Flavio Villanustre, Taghi M Khoshgoftaar, Naeem Seliya, Randall Wald, and Edin Muharemagic. 2015. Deep learning applications and challenges in big data analytics. *Journal Of Big Data* 2, 1 (2 2015). https://doi.org/10.1186/s40537-014-0007-7
[30] Erick Odhiambo Omuya, George Onyango Okeyo, and Michael Waema Kimwele. 2021. Feature Selection for Classification using Principal Component Analysis and Information Gain. *Expert Systems with Applications* 174 (2 2021), 114765. https://doi.org/10.1016/j.eswa.2021.114765
[31] Rafel Palliser-Sans. 2021. RRULES: An improvement of the RULES rule-based classifier. *arXiv (Cornell University)* (1 2021). https://doi.org/10.48550/arxiv.2106.07296
[32] Yash Paul and Neerendra Kumar. 2019. *A comparative study of famous classification techniques and data mining tools.* 627–644 pages. https://doi.org/10.1007/978-3-030-29407-6_45
[33] D.T. Pham and Aksoy. 1995. RULES: A simple rule extraction system. *Expert Systems with Applications* 8, 1 (1 1995), 59–65. https://doi.org/10.1016/s0957-4174(99)80008-6
[34] Steven L. Salzberg. 1994. C4.5: Programs for Machine Learning by J. Ross Quinlan. Morgan Kaufmann Publishers, Inc., 1993. *Machine Learning* 16, 3 (9 1994), 235–240. https://doi.org/10.1007/bf00993309
[35] Nayani Sateesh, Pasupureddy Srinivasa Rao, and Davuluri Rajya Lakshmi. 2023. Optimized ensemble learning-based student's performance prediction with weighted rough set theory enabled feature mining. *Concurrency and Computation Practice and Experience* 35, 7 (1 2023). https://doi.org/10.1002/cpe.7601

[36] Michael J. Siers and Md Zahidul Islam. 2015. Software defect prediction using a cost sensitive decision forest and voting, and a potential solution to the class imbalance problem. *Information Systems* 51 (3 2015), 62–71. https://doi.org/10.1016/j.is.2015.02.006

[37] Frederic Stahl, Mohamed Medhat Gaber, and Manuel Martin Salvador. 2012. *ERules: a modular adaptive classification rule learning algorithm for data streams*. 65–78 pages. https://doi.org/10.1007/978-1-4471-4739-8_5

[38] Chayanan Thanakiattiwibun, Arunotai Siriussawakul, Tithita Virotjarumart, Satanun Maneeon, Narisa Tantai, Varalak Srinonprasert, Onuma Chaiwat, and Patcharee Sriswasdi. 2023. Multimorbidity, healthcare utilization, and quality of life for older patients undergoing surgery: A prospective study. *Medicine* 102, 13 (3 2023), e33389. https://doi.org/10.1097/md.0000000000033389

[39] M. Thangaraj and C. R. Vijayalakshmi And. 2013. Performance Study on Rule-based Classification Techniques across Multiple Database Relations. https://www.ijais.org/archives/volume5/number4/432-0608/

[40] Elaheh Toulabinejad, Mohammad Mirsafaei, and Alireza Basiri. 2023. Supervised discretization of continuous-valued attributes for classification using RACER algorithm. *Expert Systems with Applications* 244 (8 2023), 121203. https://doi.org/10.1016/j.eswa.2023.121203

[41] Joaquin Vanschoren, Jan N. Van Rijn, Bernd Bischl, and Luis Torgo. 2014. OpenML. *ACM SIGKDD Explorations Newsletter* 15, 2 (6 2014), 49–60. https://doi.org/10.1145/2641190.2641198

[42] Caixia Wang, Xiaoyun Wei, Aiqian Yang, and Haiyan Zhang. 2022. Construction and analysis of discrete system dynamic modeling of physical education teaching mode based on decision tree algorithm. *Computational Intelligence and Neuroscience* 2022 (7 2022), 1–11. https://doi.org/10.1155/2022/2745146

[43] Dingming Wu, Xiaolong Wang, Jingyong Su, Buzhou Tang, and Shaocong Wu. 2020. A labeling method for financial time series prediction based on trends. *Entropy* 22, 10 (10 2020), 1162. https://doi.org/10.3390/e22101162

[44] Wei-Sheng Wu and Meng-Jhun Jhou. 2017. MVIAeval: a web tool for comprehensively evaluating the performance of a new missing value imputation algorithm. *BMC Bioinformatics* 18, 1 (1 2017). https://doi.org/10.1186/s12859-016-1429-3

[45] Darren Yates, Md Zahidul Islam, and Junbin Gao. 2019. *SPAARC: a fast decision tree algorithm*. 43–55 pages. https://doi.org/10.1007/978-981-13-6661-1_4